\pdfoutput=1

\documentclass[11pt]{article}
\usepackage{float}

\usepackage[table]{xcolor}
\usepackage[a-1b]{pdfx}

\usepackage{framed}
\usepackage{lmodern}
\usepackage{mdwlist}
\usepackage{siunitx}
\usepackage{latexsym}
\usepackage{colortbl}
\usepackage{nicefrac}
\usepackage{booktabs}
\usepackage{fnpct}
\usepackage{amsfonts}
\usepackage[T1]{fontenc}
\usepackage{bold-extra}
\usepackage{amsmath}
\usepackage{amssymb}
\usepackage{bm}
\usepackage{graphicx}
\usepackage{mathtools}
\usepackage{microtype}
\usepackage{multirow}
\usepackage{multicol}
\usepackage{xpatch}
\usepackage{latexsym,comment}
\usepackage[normalem]{ulem}

\newcommand*{\missingreference}{{\Huge \colorbox{red}{?reference?}}}
\newcommand*{\missingcitation}{{\Huge \colorbox{red}{?citation?}}}

\makeatletter
\xpatchcmd{\@setref}{\bfseries}{\missingreference}{}{}
\def\@citex[#1]#2{\leavevmode
    \let\@citea\@empty
    \@cite{\@for\@citeb:=#2\do
        {\@citea\def\@citea{,\penalty\@m\ }%
            \edef\@citeb{\expandafter\@firstofone\@citeb\@empty}%
            \if@filesw\immediate\write\@auxout{\string\citation{\@citeb}}\fi
            \@ifundefined{b@\@citeb}{\hbox{\reset@font\missingcitation}%
                \G@refundefinedtrue
                \@latex@warning
                {Citation `\@citeb' on page \thepage \space undefined}}%
            {\@cite@ofmt{\csname b@\@citeb\endcsname}}}}{#1}}
\makeatother

\newcommand{\gem}[1]{\mbox{\textsc{gem}}}

\newcommand{\g}{\, | \,}

\newcommand{\hidetext}[1]{}
\newcommand{\ignore}[1]{}
\newif\ifcomment
\commenttrue

\ifcomment
    \newcommand{\pinaforecomment}[3]{\colorbox{#1}{\parbox{.8\linewidth}{#2: #3}}}

    \newcommand{\prtodo}[1]{\pinaforecomment{lightblue}{pr}{#1}}
    \newcommand{\prtodoi}[1]{\pinaforecomment{lightblue}{pr}{#1}}
\else
    \newcommand{\pinaforecomment}[3]{}
    \newcommand{\prtodo}[1]{}
    \newcommand{\prtodoi}[1]{}
\fi

\newcommand{\smallurl}[1]{ \begin{tiny}\url{#1}\end{tiny}}

\definecolor{lightblue}{HTML}{3cc7ea}
\definecolor{CUgold}{HTML}{CFB87C}
\definecolor{grey}{rgb}{0.95,0.95,0.95}
\definecolor{ceil}{rgb}{0.57, 0.63, 0.81}
\definecolor{UMDred}{HTML}{ed1c24}
\definecolor{UMDyellow}{HTML}{ffc20e}

\usepackage{neurips_2024}

\usepackage{xspace}
\usepackage{pgfplots}
\usepackage{dsfont}
\DeclareUnicodeCharacter{2212}{−}
\usepgfplotslibrary{groupplots,dateplot}
\usetikzlibrary{patterns,shapes.arrows}
\pgfplotsset{compat=newest}

\usepackage{tikzscale}
\usepackage{relsize}
\usepackage{amsmath,amssymb}
\newcommand{\probP}{\text{I\kern-0.15em P}}

\usepackage{multirow, colortbl}

\usepackage{tabularx,booktabs}
\usepackage{makecell}

\usepackage[normalem]{ulem}
\useunder{\uline}{\ul}{}

\usepackage{graphicx}

\definecolor{ablation6}{HTML}{fcefed}
\definecolor{ablation_tie}{HTML}{fce3e1}

\definecolor{ablation5}{HTML}{fcd8d4}
\definecolor{ablation4}{HTML}{FBC3BC}
\definecolor{ablation3}{HTML}{F7A399}
\definecolor{ablation2}{HTML}{F38375}
\definecolor{ablation1}{HTML}{EF6351}

\usepackage[]{algpseudocode}
\usepackage[]{algorithm}
\usepackage{float}
\algtext*{EndFor}%
\algtext*{EndProcedure}%

\newcommand{\m}{\textsc{MLM}\xspace}
\newcommand{\ms}{\textsc{MLMs}\xspace}

\usepackage{booktabs}
\usepackage[normalem]{ulem}
\useunder{\uline}{\ul}{}

\usepackage{times}
\usepackage{latexsym}
\usepackage{adjustbox}

\usepackage[T1]{fontenc}
\usepackage{amsmath}
\usepackage{amssymb}
\usepackage{booktabs}
\usepackage{tikzscale}
\usepackage{amsmath}

\usepackage{multirow, colortbl}

\usepackage{tabularx,booktabs}
\usepackage{makecell}
\usepackage{multirow}
\usepackage{scalerel,xparse}

\usepackage{cleveref}
\usepackage{microtype}
\usepackage[most]{tcolorbox}

\usepackage{enumitem}
\crefformat{section}{\S#2#1#3}
\crefformat{subsection}{\S#2#1#3}
\crefformat{subsubsection}{\S#2#1#3}

\definecolor{bggray}{rgb}{0.95, 0.95, 0.95}
\usepackage[%
    framemethod=tikz,
    skipbelow=\topskip,
    skipabove=\topskip
]{mdframed}
\mdfsetup{%
    leftmargin=0pt,
    rightmargin=0pt,
    backgroundcolor=bggray,
    middlelinecolor=black,
    roundcorner=3
}

\definecolor{SkyBlue}{rgb}{0.53, 0.81, 0.92} %
\definecolor{OliveGreen}{rgb}{0.05, 0.75, 0.24}
\definecolor{BrickRed}{rgb}{0.8, 0.25, 0.33} %

\newtcolorbox[list inside=prompt,auto counter,number within=section]{prompt}[1][]{
    colbacktitle=black!60,
    fonttitle=\small,
    coltitle=white,
    fontupper=\footnotesize,
    boxsep=3pt,
    left=0pt,
    right=0pt,
    top=0pt,
    bottom=0pt,
    boxrule=1pt,
    #1
}

\definecolor{UMDred}{HTML}{ed1c24}

\definecolor{yellowcolor}{HTML}{ffc20e}
\definecolor{redcolor}{HTML}{ff8f87}
\definecolor{bluecolor}{HTML}{8cd2f5}

\title{
Can They Dixit? Yes they Can!\\
Dixit as a Playground for
Multimodal Language Model Capabilities}

\author{
    Nishant Balepur \\
    University of Maryland \\
    \texttt{nbalepur@umd.edu} \\\And
    Dang Nguyen \\
    University of Maryland \\
    \texttt{dangmn@umd.edu} \\\And
    Dayeon Ki \\
    University of Maryland \\
    \texttt{dayeonki@umd.edu}
}

\begin{document}

\maketitle

\begin{abstract}
Multi-modal large language models (\ms) are often assessed on static, individual benchmarks---which cannot jointly assess \m capabilities in a single task---or rely on human or model pairwise comparisons---which is highly subjective, expensive, and allows models to exploit superficial shortcuts (e.g., verbosity) to inflate their win-rates.
To overcome these issues, we propose game-based evaluations to holistically assess \m capabilities.
Games require multiple abilities for players to win, are inherently competitive, and are governed by fix, objective rules, and makes evaluation more engaging, providing a robust framework to address the aforementioned challenges.
We manifest this evaluation specifically through Dixit, a fantasy card game where players must generate captions for a card that trick some, but not all players, into selecting the played card.
Our quantitative experiments with five \ms show Dixit win-rate rankings are perfectly correlated with those on popular \m benchmarks, while games between human and \m players in Dixit reveal several differences between agent strategies and areas of improvement for \m reasoning. \footnote{\url{https://github.com/nbalepur/dixit}}
\end{abstract}

\section{Introduction}

Multi-modal large language models (\ms) have made significant progress, showing impressive performance in tasks requiring both image and text inputs, such as image captioning, classification, and visual understanding \citep{zheng2023steve}.
As a testament to this, several \ms with diverse architectures have been introduced in the past few years, each aiming to excel in such tasks \citep{chen2024internvl, bai2023qwen, agrawal2024pixtral}.
As the number of \ms continues to grow, the need for robust and comprehensive evaluation frameworks becomes increasingly important \citep{white2024livebench, saxon2024benchmarks}.
These frameworks are key for reliably comparing models, enabling users to choose those that best align with their needs \citep{li2024survey}.

Such \m benchmarks fall into two main categories.
The first approach evaluates models on individual tasks---such as image captioning, classification, or understanding---pinpointing abilities in isolation \citep{duan2024vlmevalkit, lee2024vhelm}. 
While useful, this precludes jointly testing model capabilities allowing for direct comparisons between models. 
The second involves sampling outputs from two models and using pairwise comparison judgments from humans or other models to identify which model is better (i.e. win-rate) \citep{ChatArena, chiang2024chatbot}.
While this better compares models, such evaluations are inherently subjective \cite{lopez2021preference}, as they rely on external human or model judgments without objective rules \citep{zheng2023judging}, and such judges are prone to biases like verbosity and position \citep{wang2023large, shi2024judging, ye2024justice}.
They are also expensive, requiring large-scale human annotation or computational resources for model-based judging \citep{lee2023rlaif, chen2024humans}.

\begin{figure*}
    \centering
    \includegraphics[width=\linewidth]{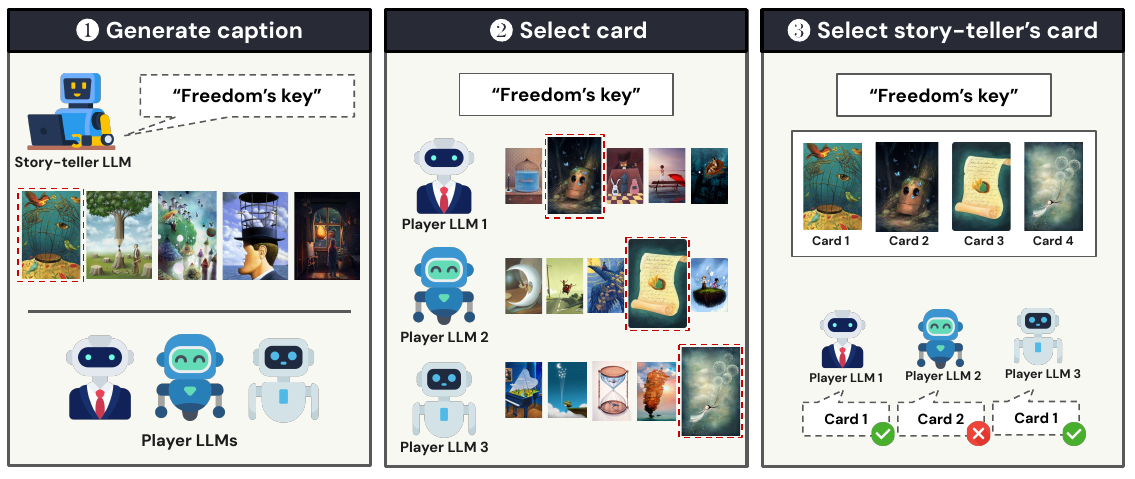}
    \caption{Overview of playing Dixit with MLMs. \textbf{Step 1}: The story-teller model generates a caption ``\textit{Freedom's key}'' for one of its cards, which is shown to all players. \textbf{Step 2}: Each player model select a card from their hand that they think corresponds to the caption. \textbf{Step 3}: From the pool of selected cards, all player models select the card they believe was the story-teller's.}
    \label{fig:main_figure}
\end{figure*}

To address these issues, we draw from Natural Language Processing (NLP) research showing multi-agent LLM evaluations can be conducted through games \citep{xu2024survey, ye2024justice, hu2024survey}, and propose game-based evaluations as a framework for evaluating \ms.
Games inherently involve competition \citep{lebed2006system}, enabling direct comparison of model capabilities, with rankings serving as a proxy for model performance.
Further, games are governed by fixed rules \citep{devries2015games}, eliminating the need for external judges and ensuring objective, efficient evaluations.
The rules of games are also designed in a manner to prevent players, in our case models, from exploiting superficial shortcuts to inflate their win-rates \citep{webb2007cheating}.
As an added benefit, games are designed to be fun \citep{prensky2001fun}, so practitioners can directly intervene in game-based evaluations to assess their models in a more engaging manner.

To advocate for this approach, we adopt Dixit \citep{kunda2020creative, hsu2019dixit}---a fantasy card game---as our game-based evaluation suite for \ms, as it evaluates several diverse \m capabilities (Figure~\ref{fig:main_figure}).
In Dixit, one player called the \textit{storyteller} selects a card from their hand and generates a caption for it.
The other players then choose cards from their hands that they believe best match the caption.
All of these cards are pooled, and players vote on which card (not their own) they think is the storyteller's.
Non-story teller players score points if they select the correct card, but the storyteller faces a unique challenge to score points: their caption must be sufficiently relevant to guide other players to the correct card but not so obvious such that all players guess correctly; the storyteller does not score points if either no one or everyone guesses their card.

Dixit provides a unified and nuanced evaluation suite for multiple \m capabilities.
A strong player must be adept at: \textbf{1)} generating creative captions that allow only some users to pick the correct card (testing creativity, image comprehension, and calibration in image captioning); and \textbf{2)} identifying cards that align with what players believe the storyteller to have played and what players believe other players may select (testing accuracy, image comprehension, and theory-of-mind in image classification).
Dixit captures diverse model abilities in a single, cohesive task, making it an appealing game for our analysis.

We build our own evaluation interface for Dixit and have five \ms compete as players: GPT-4o \citep{gpt4}, Claude-3.5 Sonnet \citep{anthropic2023claude}, Intern-VL2 \citep{chen2024internvl}, Qwen2VL \citep{bai2023qwen}, and Molmo \citep{deitke2024molmo}.
We find that all players significantly outperform a player who randomly generates captions and selects cards, showing that \ms that are not directly trained on Dixit can still demonstrate above-random Dixit capabilities.
Further, ranking players by Dixit performance perfectly matches the rankings on popular leaderboards like ChatbotArena \citep{chiang2024chatbot} and the Open VLM Leaderboard \citep{duan2024vlmevalkit}, suggesting that Dixit comprehensively tests \m capabilities within a single task.
We then have humans play against one \m to study if \ms can surpass humans in Dixit.
GPT-4o Mini, a relatively weak \m, was able to surpass one human player, showing that \ms possess strong Dixit capabilities that rival and may even surpass human abilities.
Finally, we assess the game logs between \m players in Dixit to understand differences between model and human Dixit strategies; we find when humans are story-tellers, they tend to generate more abstract captions that reference external knowledge, while \m captions are easier as they literally describe the input image, suggesting future models can improve in Dixit by enhancing their commonsense reasoning capabilities.

We advocate for game-based frameworks, like Dixit, to evaluate \m capabilities.
Such games allow us to jointly compare several model abilities in a single unified task without relying on expensive and biased external judges.
It also allows us to make evaluations more engaging, as humans can directly intervene in games as players to evaluate their models.

\section{Related Work}

\subsection{Playing Games with \ms}
Multimodal LLMs (\ms) have demonstrated their potential in tasks involving text and image input, such as image captioning and classification \citep{zheng2023steve}. Beyond these conventional tasks, \ms have been increasingly applied as agents to perform complex tasks, where they have exhibited impressive generalization capabilities in dynamic environments \citep{tan2024trueknowledgecomespractice, wang2024describe}. This progress has sparked significant interest in the use of \ms in game playing, a domain that naturally integrates reasoning, strategy, and multimodal understanding.

Previously, \ms have been applied to popular games like Minecraft \citep{lifshitz2024steve} to navigate and interact within intricate environments and as low-level controllers in Atari video games to assess their ability to interpret complex visual scenes and formulate strategies \citep{waytowich2024atarigptbenchmarkingmultimodallarge}. Similarly, game-based benchmarks like MatchIt \citep{chalamalasetti2023clembench, hakimov2024usinggameplayinvestigate} require \ms to communicate and compare visual input, which assess \ms's abilities in effective communication and image reasoning.

Similarly, we adopt \ms in playing games, but uniquely focuses on the game Dixit, which has yet to be explored in the context of \ms. Although Dixit has been studied in psychology, education, and cognitive science, its use as a benchmark for multimodal reasoning remains underexplored \citep{kunda2020creativecaptioningaigrand}.
Finally, a few works have explored \m captioning capabilities in Dixit \citep{hsu2019dixit}, but we are the first to have \m agents compete in Dixit end-to-end in a full game to holistically evaluate their capabilities.

\subsection{Evaluation of \ms}
Evaluation of vision-language tasks typically employs two main approaches: \textbf{1)} pointwise \citep{liu2024visual, yu2023mm, sun2023aligning, zhang2025mathverse} and \textbf{2)} pairwise settings. Pointwise evaluations focus on isolated tasks such as image captioning \citep{nguyen2023improvingmultimodaldatasetsimage}, classification \citep{abdelhamed2024seeenhancingzeroshotimage}, or visual understanding \citep{wang2024advancedmultimodaldeeplearning}. While effective for assessing specific skills, this approach is limited in its ability to jointly test and compare models' broader multimodal capabilities and often introduces subjectivity in interpreting results. On the other hand, pairwise evaluations, where outputs from two models are sampled and compared directly \citep{lu2024wildvision, yu2024rlaif}, address some of these limitations but are resource-intensive, requiring significant time, computational power, and human effort. In response to these, we propose a game-based evaluation framework that simultaneously tests multiple capabilities within a structured, rule-governed environment.

\begin{figure*}
    \centering
    \includegraphics[width=\linewidth]{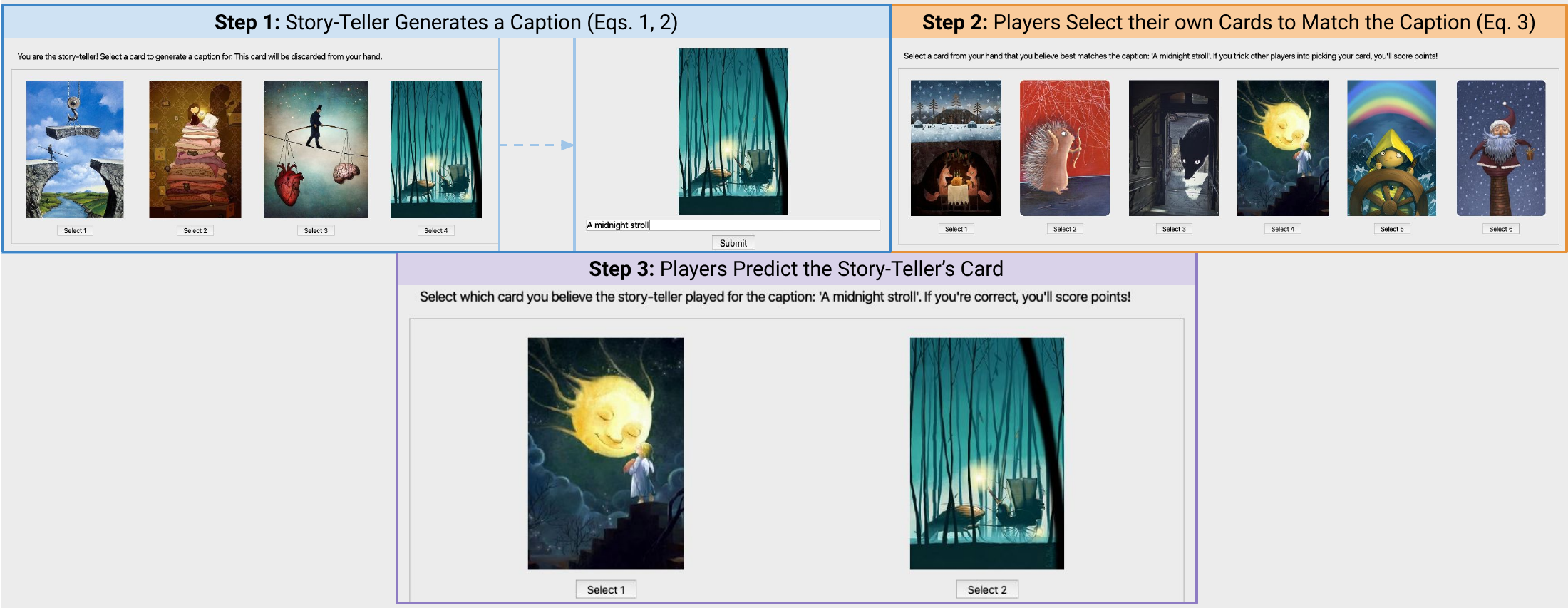}
    \caption{Visualization of our graphical user interface for humans to play Dixit. Human players can generate captions for cards and select cards from their hand, just like in the original game of Dixit.}
    \label{fig:human}
\end{figure*}

\section{Method: The Rules of Dixit}

Each game of Dixit uses a deck of $m$ cards $\mathcal{C} = \{c_1, c_2, ..., c_m\}$ and consists of $3 \leq n \leq 6$ players $\mathcal{P} = \{p_1, p_2, ..., p_n\}$.
Each card $c \in \mathcal{C}$ contains a fantasy picture (Figure~\ref{fig:main_figure}) to allow players to exercise creativity in caption generation.
Each player $p_i \in \mathcal{P}$ draws $n$ cards to form their hand $\mathcal{C}_i \subset \mathcal{C}$ and $\mathcal{C}_i \cap \mathcal{C}_j = \emptyset \; \; \forall p_i, p_j \in \mathcal{P}$ for $p_i \neq p_j$ (i.e. each card is unique). 
In a single round of Dixit, one player $p_j \in \mathcal{P}$ is the \textit{story-teller}; the story-teller role cycles between players in each round.
After every round, each player draws an unseen card from $\mathcal{C}$ to replace the card they played in the round.
The game ends when any player scores 30 or more points, or if the deck $\mathcal{C}$ is out of unseen cards.
Below, we describe each step of Dixit and the scoring system (\cref{subsection:score}).

\subsection*{Step 1: The Story-Teller Generates a Caption.} \label{subsection:step1}

In the first step of each turn in Dixit, the story-teller $p_j \in \mathcal{P}$ must generate a caption $t$ for one of the cards in their hand $c \in \mathcal{C}_j$.
We decompose this into two steps.
First, using their hand $\mathcal{C}_j$ as input, $p_j$ must sample a card $c \sim p_j(\mathcal{C}_j)\in \mathcal{C}_j$ to generate a caption for. Next, using the story-teller's selected card $c$ as input, $p_j$ must generate a caption $t \sim p_j(c)$ for this card. For the story-teller to score points, the caption $t$ should allow some of the other players $p_i \in \mathcal{P} \setminus \{p_j\}$ to identify the correct card $c$, but not all of them.
As a result, when humans play Dixit, captions are often vague.
This poses a unique challenge beyond traditional image captioning tasks, where the goal is to produce a highly-accurate caption for an input image.

\subsection*{Step 2: Players Select Cards from their Own Hand for the Caption.}
\label{subsection:step2}

After the story-teller generates a caption $t$, it is read aloud or shown to all players $\mathcal{P}$.
After understanding the caption, each player except the story-teller $p_i \in \mathcal{P} \setminus \{p_j\}$ must select a card from their hand $c_{i} \sim p_i(t, \mathcal{C}_i) \in \mathcal{C}_i$ that they believe corresponds to the caption $t$. Players score points in the third step of Dixit (\cref{subsection:step3}) if other players incorrectly pick their card when they guess which is the story-teller's, so each player is incentivized to select the card from their hand that they believe best corresponds to the caption $t_j$.

After all non-story-teller players select cards from their hands, these cards are combined with the story-teller's card to create a pool of cards $\mathcal{C}_{pool} = \{ c \} \cup \{c_i \g \forall p_i \in \mathcal{P} \setminus \{p_j\} \}$. Pooling cards also ensures that the story-teller does not generate a fully random caption $t_j$ with the goal of having just one other player selecting this card in the third step (\cref{subsection:step3}) by chance; if the caption is random and another player has a card that better matches it, most players will pick this relevant card and the story-teller will not score any points (\cref{subsection:score}).

\subsection*{Step 3: Players Select the Card they Believe is the Story-Teller's.} 
\label{subsection:step3}

After the pool of cards $\mathcal{C}_{pool}$ is created, all non-story-teller players $p_x \in \mathcal{P} \setminus \{p_i\}$ must select the card $\hat{c}_x \sim p_x(t, \mathcal{C}_{pool} \setminus \{c_x\}) \in \mathcal{C}_{pool}$ they believe was the story-teller's card $c_j$ used to create the caption $t_j$. Each player $p_i$ knows that their own card $c_x$ is not the story-teller's card $c$, so for simplicity, we remove this card from $\mathcal{C}_{pool}$ when giving each player the pool of cards they can select from.

\subsection*{Scoring System} 
\label{subsection:score}

After each turn, the player's votes are tallied, and scoring is calculated as follows.
The story-teller receives: 0 points if \textit{no} player picked the story-teller's card, i.e., $\hat{c}_i \neq c ;\ \forall p_i \in \mathcal{P} \setminus \{ p_j \}$; 0 points if \textit{all} players picked the story-teller's card, i.e., $\hat{c}_i = c ;\ \forall p_i \in \mathcal{P} \setminus \{ p_j \}$; 3 points otherwise. Each non-story-teller receives: 2 points if the story-teller received 0 points; 3 points if the story-teller did not get 0 points and the player picked the story-teller's card; 0 points otherwise.
Non-story-tellers also receive an additional +1 point for every player who incorrectly believed their card was the story-teller's (i.e. player $p_i$ gets additional points equal to $\sum_{x} \mathrm{I}(\hat{c}_x = c_i)$).

\section{Experimental Setup}

\begin{table*}[t]
\small
\centering
\setlength{\tabcolsep}{4pt}
\begin{tabular}{@{}ccc|ccc@{}}
\toprule
Model             & Avg Points ($\uparrow$) & Avg Position ($\downarrow$) & Dixit Rank ($\downarrow$) & OpenVLM Rank ($\downarrow$) & \begin{tabular}[c]{@{}c@{}}ChatArena Rank ($\downarrow$)\end{tabular} \\ \midrule
GPT-4o            & \textbf{29.25}       & \textbf{1.725}               & \textbf{1}                   & \textbf{1 (4)}                              & \textbf{1 (3)}                                                                                 \\
Claude-3.5 & 28.55                & 2.050                        & 2                            & 2 (7)                                       & 2 (7)                                                                                 \\
Qwen-2-VL         & 25.25               & 3.075                        & 3                            & 3 (18)                                      & 3 (19)                                                                                \\

InternVL2         & 22.70                & 3.725                        & 4                            & 4 (30)                                      & 4 (21)                                                                                \\

Molmo             & 18.80                & 4.525                        & 5                            & 5 (73)                                      & 5 (25)                                                                                \\
Random            & 8.85                 & 5.900                        & 6                            & -                                           & -                                                                                     \\ \bottomrule
\end{tabular}
\vspace{1ex}
\caption{Benchmarking of MLM Dixit capabilities (left) and comparing MLM Dixit rankings versus rankings on popular MLM benchmarks (right). The \textit{best} performing model is in \textbf{bold}.}
\label{table:model_arena}
\end{table*}

\subsection{Dataset}

To have \ms play Dixit, we need images of Dixit cards that we can use as inputs.
The Dixit game is proprietary and subject to copyright laws, so we cannot use images from the actual game.
Thus, we use cards from an open-source version of Dixit\footnote{\url{https://github.com/jminuscula/dixit-online}}, where cards are obtained from Pinterest, ensuring that we use our data within its intended use.
In total, we obtain $m = 100$ cards to form the deck $\mathcal{D}$.

\subsection{Models}

We evaluate five strong \ms in Dixit consisting of two proprietary models (GPT-4o \citep{gpt4} and Claude 3.5 Sonnet \citep{anthropic2023claude}), and three open-weights models (InternVL2-8B \citep{chen2024internvl}, Molmo-7B-D \citep{deitke2024molmo}, Qwen2-VL-7B \citep{bai2023qwen}).\footnote{Model version or HuggingFace names are outlined in Appendix Table \ref{tab:models}.}

We also implement a trivial Random player to ensure these models have above-random Dixit capabilities; when the Random player is the story-teller, it selects a caption randomly from a list of 30 captions generated by GPT-4o, and when selecting cards, it randomly samples a card using a Uniform distribution. All models are zero-shot prompted and asked to provide the rationales for their decisions. Exact prompts are shown in Appendix \ref{sec:prompts}.

\subsection{Human Interface}

To allow humans to intervene in our Dixit-based evaluation and compare human and model abilities in the game, we also implement a graphical user interface (GUI) for users to play Dixit (Figure~\ref{fig:human}). 
Our GUI is implemented with the \text{PyQt5} library.\footnote{\url{https://pypi.org/project/PyQt5/}}

\begin{table*}[t]
\small
\centering
\setlength{\tabcolsep}{4.5pt}
\begin{tabular}{@{}ccccccccc@{}}
\multicolumn{1}{l}{}             & \multicolumn{2}{c}{\textit{Game 1}}                            & \multicolumn{2}{c}{\textit{Game 2}}                            & \multicolumn{2}{c}{\textit{Game 3}}                            & \multicolumn{2}{c}{\textit{Average}}      \\ \toprule
\multicolumn{1}{c|}{Player}      & Points ($\uparrow$) & \multicolumn{1}{c|}{Rank ($\downarrow$)} & Points ($\uparrow$) & \multicolumn{1}{c|}{Rank ($\downarrow$)} & Points ($\uparrow$) & \multicolumn{1}{c|}{Rank ($\downarrow$)} & Points ($\uparrow$) & Rank ($\downarrow$) \\ \midrule
\multicolumn{1}{c|}{Player 1}     & 31.00               & \multicolumn{1}{c|}{1}                   & 31.00               & \multicolumn{1}{c|}{1}                   & 29.00               & \multicolumn{1}{c|}{2}                   & 30.33               & 1.33                \\
\multicolumn{1}{c|}{Player 2}        & 28.00               & \multicolumn{1}{c|}{2}                   & \textbf{19.00}               & \multicolumn{1}{c|}{\textbf{4}}                   & 30.00               & \multicolumn{1}{c|}{1}                   & 25.67               & 2.33                \\
\multicolumn{1}{c|}{Player 3}        & 24.00               & \multicolumn{1}{c|}{3}                   & 23.00               & \multicolumn{1}{c|}{2}                   & \textbf{14.00}      & \multicolumn{1}{c|}{\textbf{4}}          & \textbf{20.33}      & 3.00                \\
\multicolumn{1}{c|}{GPT-4o Mini} & \textbf{23.00}      & \multicolumn{1}{c|}{\textbf{4}}          & 22.00      & \multicolumn{1}{c|}{3}          & 21.00               & \multicolumn{1}{c|}{3}                   & 22.00               & \textbf{3.33}       \\ \bottomrule
\end{tabular}
\vspace{1ex}
\caption{GPT-4o mini vs. three authors in Dixit over three games. The \textit{worst} player is in \textbf{bold}.} \label{table:human}
\end{table*}

\section{Results}

\subsection{Which \m is the Strongest Dixit Player?}

We first understand the relationship between typically measured \m capabilities and those measured in Dixit by having our six \ms play 20 games of Dixit.
To measure overall \m performance on Dixit, we use two metrics: \textbf{1) Average Points:} the total number of points scored by each \m divided by the number of games played (20); and \textbf{2) Average Position:} the game position of each \m (e.g. 1 for first, 2 for second) averaged over all games played (20).
The player who starts as the story-teller in each game is randomized.

Our games of Dixit with \ms reveal two notable findings (Table~\ref{table:model_arena}).
First, all \ms greatly outperform the Random player, showcasing above-random Dixit capabilities. \ms are unlikely to have been trained on Dixit logs, meaning that an \ms acquired skills in image comprehension, reasoning, and classification from other tasks are transferrable to Dixit---a likely out-of-domain task. 
Second, Dixit game rankings are perfectly correlated with rankings on OpenVLM and ChatbotArena.
As a result, we believe game-based evaluations have the potential to evaluate \ms as holistically as multi-task benchmarks.
Dixit assesses several \m capabilities---such as image comprehension, reasoning, classification, and captioning---within a single task, eliminating the need for practitioners to use several diverse task formats, and datasets to comprehensively assess \m capabilities.

\begin{figure*}
    \centering
    \includegraphics[width=\linewidth]{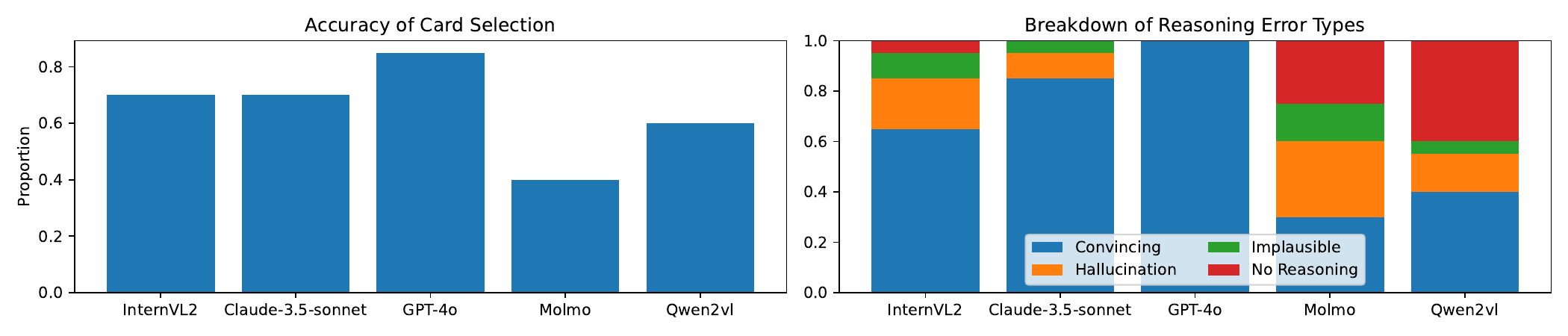}
    \caption{Accuracy of MLM selections for story-teller captions (left) along with an error analysis of the rationales provided for the selections (right) in each model's generated chain-of-thought.}
    \label{fig:accuracy}
\end{figure*}

\begin{table}[t]
\setlength{\tabcolsep}{3pt}
\centering
\small
\begin{tabular}{@{}c|cccc@{}}
\toprule
\textbf{Player} & Player 1       & Player 2 & Player 3 & GPT-4o Mini \\ \midrule
Player 1         & 1.00          & -    & -    & -           \\
Player 2            & \textbf{0.61} & 1.00 & -    & -           \\
Player 3            & 0.35          & 0.53 & 1.00 & -           \\
GPT-4o Mini     & 0.53          & 0.47 & 0.56 & 1.00        \\ \bottomrule
\end{tabular}
\caption{\label{table:agreement} Raw agreement between human and Dixit players. Highest non-perfect agreement is bold.}
\end{table}

\subsection{Are Humans Dominated in Dixit?}

To study if \ms have the potential to surpass humans in playing Dixit, our three authors play Dixit with an \m.
As we did not have the resources to host our human Dixit interface online, we run it locally, precluding the use of open-weight \ms.
Due to API cost constraints, we selected GPT-4o Mini for this analysis.
Since GPT-4o Mini is the weakest \m out of the ones we studied, if the model is able to beat any human, it provides evidence that \ms can play Dixit on par with humans.
In future iterations of this work, we plan to extend our human versus \m analysis to more capable and a wider variety of models.

We run three practice rounds of Dixit so the human players are confident on how to play the game.
We play three games of Dixit, each lasting around 30 minutes.
All players verbally expressed that the games were enjoyable (average Likert rating of 4.433), showing that Dixit and game-based evaluations could be a more engaging method to allow practitioners to directly intervene in their evaluations.
As engagement is often linked to education, it would be interesting for future work to see if game-based evaluations allow model designers to learn more about their models \citep{oblinger2004next}.

We show the results from our games in Table~\ref{table:human}.
Humans tend to have the upper-hand on \ms, measured through the points in each game and the rank of the players.
However, GPT-4o Mini was able to beat one player (Player 3) in points on average.
Since this model is our weakest closed-source model, we speculate that \ms could demonstrate Dixit capabilities that are similar to or maybe even surpass humans.

\subsection{What Strategies do \ms Use in Dixit?}

For a fine-grained analysis into the types of strategies \ms employ in Dixit, we study the logs and interactions between \ms when they play Dixit.
We decompose our analysis into three questions.

\subsubsection{How Accurate are \m Story-Teller Selections?}

\noindent \textbf{\m Card Selection Accuracy:} To assess \m image comprehension and reasoning capabilities in Dixit, we study the accuracy of \m decisions when they are prompted to select the story-teller's card. We first have one Ph.D. student annotate if the card selected by the \m plausibly matches the caption of the story-teller, which we call \textbf{accuracy}. 

In Figure~\ref{fig:accuracy} (left), we find that most \ms have similar, moderate accuracy, around 0.70. The notable exception in GPT-4o, which has an accuracy over 0.80; we believe this improvement over other \ms can explain the model's high win rate when models compete against each other.

\noindent \textbf{\m Reasoning Errors:} Since all \ms are prompted to select cards via chain-of-thought, we analyze the rationales provided by the models to identify any potential reasoning gaps that could be improved upon in future work. We classify rationales into four types: \textbf{1) Convincing:} No issues; \textbf{2) Implausible:} The rationale behind the selected card is unrealistic or far-fetched; \textbf{3) Hallucination:} The rationale references something not present in the card; and \textbf{4) No Reasoning:} The model does not engage in reasoning (e.g. simply repeats the caption).

\begin{figure*}
    \centering
    \includegraphics[width=\linewidth]{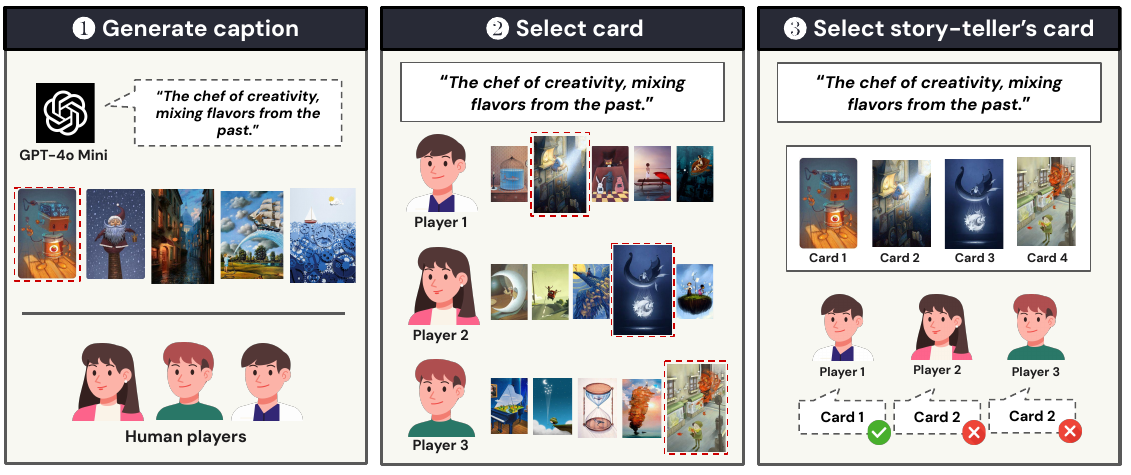}
    \caption{Illustration of example when GPT-4o Mini generates a caption that fooled 2/3 human players.}
    \label{fig:case_study}
\end{figure*}

In Figure~\ref{fig:accuracy} (right), we find that some models, in particular closed-source models, are able to provide convincing justifications for their decisions via chain-of-thought; open-source models lag behind, revealing that model practicioners can aim to design models with improved chain-of-thought reasoning. Further, we find that the majority of errors are from models either refusing to engage in reasoning or hallucinating items that are not present in the image. Thus, Dixit could serve as a valuable testbed to study \m hallucinations and reasoning \citep{wu2024autohallusion}.

\noindent \textbf{\m and Human Card Selection Correlation:} We now explore the raw agreement between different human players and GPT-4o Mini from our human versus \m games when players must select the story-teller's card from the same set of cards. In Table~\ref{table:agreement}, we find that the highest agreement are between two human players: Player 1 and Player 2. However, GPT-4o Mini also has fairly high agreement with some human players, always exceeding 0.50. 
This highlights that Dixit is a highly subjectively game, making it a valuable testbed for creativity and subjective decision-making.
Further, it shows GPT-4o Mini aligns with human reasoning patterns reasonably well in the Dixit image classification tasks.

\subsubsection{What Types of Captions do \ms Generate?}
\begin{figure}
    \centering
    \includegraphics[width=\linewidth]{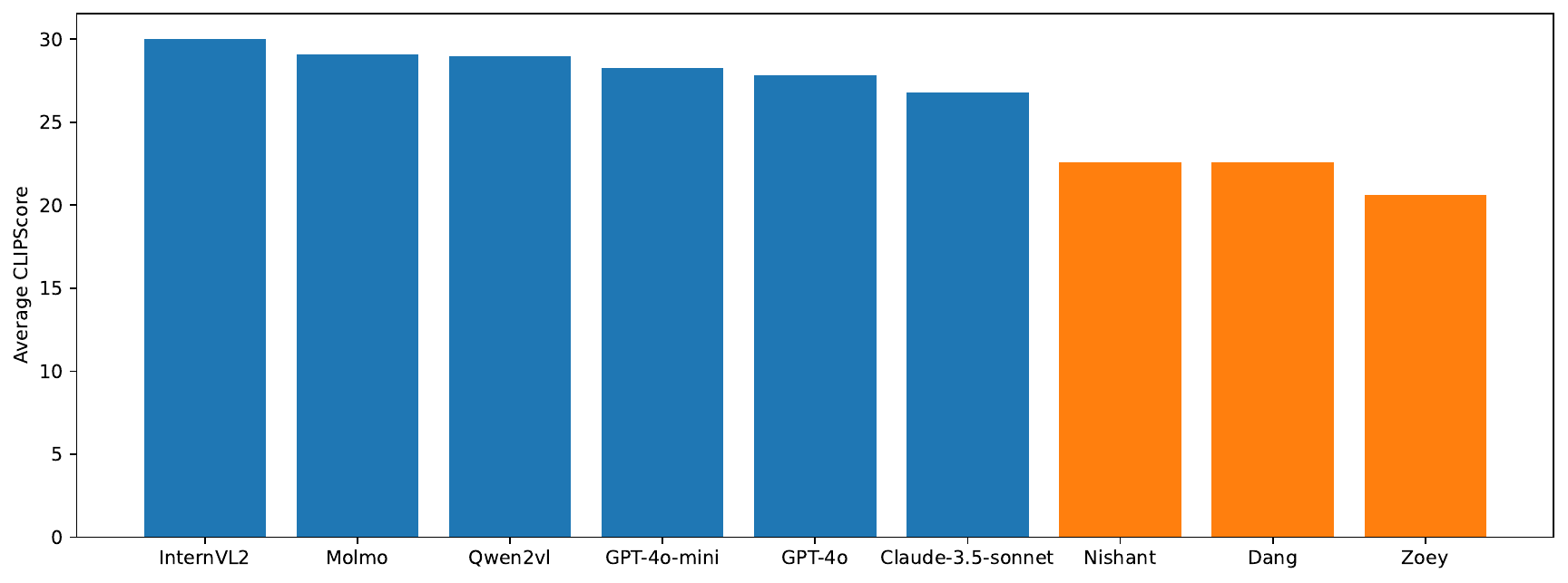}
    \caption{Average CLIPScore between images and generated captions among different players. Orange columns denote human players.}
    \label{fig:clip_scores}
\end{figure}

In this section, we aim to understand the differences between the captioning strategies of humans and \ms. 
We compute the average number of tokens in \m and human-generated captions; \m captions have 11.02 tokens on average, while human ones have 3.57 tokens, showing \ms tend to generate lengthier, more descriptive captions compared to humans.

Further, upon manual inspection, we find \ms tend to engage in \textit{literal} Dixit strategies---describing exactly what is occurring in the picture (e.g. ``Child reaching for the moon'', ``A forgotten journey through enchanted woods'').
In contrast, humans tend to reference external knowledge and pop culture when generating captions (e.g. ``Rapunzel'', ``Marceline from Adventure Time''); we speculate such references are more fun for players and harder in the image selection stage, as it requires an extra hop of commonsense reasoning \citep{bauer2018commonsense}.
As a result, we believe future work in designing optimal \m Dixit players can find ways to instill external commonsense knowledge while playing \cite{wang2020visual}. 

To confirm these differences in strategies quantitatively, we employ CLIPScore \citep{hessel-etal-2021-clipscore}, which measures the semantic similarity between each image and its corresponding generated caption.
We compute the average CLIPScores of captions generated by \ms and humans in Figure~\ref{fig:clip_scores}.
Our results align with our qualitative assessment: humans tend to generate more ambiguous captions, allowing for various interpretations, while \ms often produce captions that precisely describe the given images, making it easier for other players to identify the correct card.

\subsubsection{When Do \ms Generate Captions that Fool Most Human Players?}

We analyze a case where GPT-4o Mini served as the storyteller and successfully generated a caption that fooled most human players in the human versus MLM game, as shown in Figure \ref{fig:case_study}. The caption generated by GPT-4o Mini for its selected card was, ``\textit{A chef of creativity, mixing flavors from the past.}'' The reasoning behind this caption was that the quirky character depicted in the card image appeared ready to ``serve memories from a can.'' Each non-storyteller human player then selected a card from their hand to contribute to the pool of options. Among the pool of cards, only one human player (Player 1) correctly identified the storyteller's card, while the other two human players (Player 2 and Player 3) mistakenly selected Player 1's card. GPT-4o Mini's success in fooling the human players highlights its capability to craft a caption that is not \textit{literal} but rather ambiguous, allowing for multiple interpretations. This made it challenging for the human players to identify the story-teller's card.

\section{Conclusion}

In this paper, we explore the utility of Dixit as a testbed for holistically evaluating \m capabilities, including: image comprehension, image captioning, and image classification.
While \ms excel in these capabilities, demonstrated by above-random Dixit performance in our offline analysis and competitive win-rates that rival human players, we also uncover several areas of improvement for future Dixit \m models; notably, open-source \m showcase significant reasoning weaknesses in image classification---prone to hallucinations and lack step-by-step breakdowns, while all models tend to generate detailed and literal captions for input images---unlike humans who often produce shorter caption that reference external knowledge and commonsense.
We encourage future work to explore training strategies beyond basic prompting techniques to build stronger Dixit models.
Specifically, we believe it will be an interesting challenge to design models that not only excel in Dixit win-rate, which can be achieved via strategies like self-play \cite{silver2017mastering}, but also make the game enjoyable for other human players.
Overall, we argue that game-based evaluations like Dixit are a robust, challenging, and engaging testbed for \m capabilities, and we encourage future works to explore these setups further to evaluate their models within a single, unified task.

\section{Limitations}
We selected Dixit as our testbed for the several reasons: (1) Dixit tests MLM's creative captioning and calibration capability that only allow limited number of users to pick the correct card. (2) Dixit tests MLM's theory-of-mind in image classification since they need to pick cards in consideration of the other players' strategies. To this end, Dixit provides a unified evaluation test for testing multiple MLM capabilities. Despite these strengths, our findings should be interpreted with the limited scope of this task in mind. Nevertheless, we will make the card images, code, and framework publicly available so that new strategies, as well as tasks, can be seamlessly added.

Dixit is a game that inherently reflects the shared or non-shared knowledge of its players when generating captions. For instance, when playing Dixit with friends, one might strategically use non-shared knowledge to make it harder for others to select the correct card or rely on shared knowledge to make the game more entertaining. Consequently, a MLM's performance in playing Dixit is likely influenced by its inherent training distribution and alignment process. Our current setup does not fully account for this, leaving open questions about how model family characteristics influence performance. Future research could explore this by comparing models within the same family to determine whether shared training distributions contribute to improved performance.

\bibliography{neurips_2024}
\bibliographystyle{acl_natbib}

\clearpage

\appendix
\section{Prompts} \label{sec:prompts}

\begin{table}
\centering
\resizebox{\linewidth}{!}{%
    \begin{tabular}{llllllll}
    \specialrule{1.3pt}{0pt}{0pt}
    \textbf{Model} & \textbf{Name} \\
    \toprule

    \textbf{GPT-4o} & \texttt{gpt-4o-2024-08-06} \\
    \textbf{Claude 3.5 Sonnet} & \texttt{claude-3-5-sonnet-20241022} \\
    \textbf{InternVL2-8B} & \texttt{InternVL2-8B} \\
    \textbf{Molmo-7B-D} & \texttt{Molmo-7B-D-0924} \\
    \textbf{Qwen2-VL-7B} & \texttt{Qwen2-VL-7B-Instruct} \\
    
    \specialrule{1.3pt}{0pt}{0pt}
    \end{tabular}
}
\caption{Model version or HuggingFace model names of tested \ms.} 
\label{tab:models}
\end{table}

\lstset{
    breaklines=true,         %
    breakatwhitespace=true,  %
    basicstyle=\ttfamily,    %
    columns=fullflexible,    %
    backgroundcolor=\color{gray!10} %
}

\begin{prompt}[title={Prompt \thetcbcounter: Dixit Rules Explanation}, label=prompt:step0]
\scriptsize 
\textbf{\# Dixit Rules} \\
Dixit is a game of storytelling and imagination. Each round, one player takes on the role of the storyteller. The storyteller picks one of the six images in their hand and thinks of a sentence or phrase that describes that image. The other players then select an image from their hand that best matches the sentence given by the storyteller. The chosen images are shuffled and revealed, and players must guess which image belongs to the storyteller. Points are awarded to the storyteller if some, but not all, players guess the correct image. Points are also awarded to players whose images are selected. The game ends when the deck of cards is empty, and the player with the most points wins. \\

\textbf{\# Scoring} \\
- If all players guess the storyteller's image, the storyteller receives no points, and all other players receive 2 points. \\
- If no players guess the storyteller's image, the storyteller receives no points, and all other players receive 2 points (+1 bonus point for each vote their image receives). \\
- If some, but not all, players guess the storyteller's image, the storyteller receives 3 points, and each player who guessed correctly receives 3 points (+1 bonus point for each vote their image receives).
\end{prompt}

\begin{prompt}[title={Prompt \thetcbcounter: Selecting a Story-Teller Card}, label=prompt:step1]
\scriptsize 
\texttt{\{\{dixit\_rules\}\}} \\

\textbf{\# Instructions} \\
You are the storyteller for this round, and you have selected the following image from your hand. Think of a sentence or phrase that describes this image. Your output must follows the following JSON format:
\begin{lstlisting}
```json
{
    "thought": "Your thought here.",
    "caption": "Your caption here."
}
```
\end{lstlisting}
\end{prompt}

\begin{prompt}[title={Prompt \thetcbcounter: Generating a Story-Teller Caption}, label=prompt:step2]
\scriptsize 
\texttt{\{\{dixit\_rules\}\}} \\

\textbf{\# Instructions} \\
You are the storyteller for this round, and the images below are the cards in your hand. Select an image to generate a caption for. Your output must follows the following JSON format:
\begin{lstlisting}
```json
{
    "thought": "Your thought here.",
    "choice": "Write a single integer representing the index of the card you selected. Valid choices are {{valid_choices}}.",
}
```
\end{lstlisting}
\end{prompt}

\begin{prompt}[title={Prompt \thetcbcounter: Selecting a Card for the Pool}, label=prompt:step3]
\scriptsize 
\texttt{\{\{dixit\_rules\}\}} \\

\textbf{\# Instructions}\\
You are a non-storyteller player for this round. The storyteller provided the following caption: \texttt{\{\{caption\}\}}. The images below are the cards on the table. Select an image that you think belongs to the storyteller. Your output must follows the following JSON format:
\begin{lstlisting}
```json
{
    "thought": "Your thought here.",
    "choice": "Write a single integer representing the index of the card you selected. Valid choices are {{valid_choices}}.",
}
```
\end{lstlisting}
\end{prompt}

\begin{prompt}[title={Prompt \thetcbcounter: Voting on the Story-Teller's Card}, label=prompt:step4]
\scriptsize 
\texttt{\{\{dixit\_rules\}\}} \\

\textbf{\# Instructions} \\
You are a non-storyteller player for this round. The storyteller provided the following caption: \texttt{\{\{caption\}\}}. The images below are the cards in your hand. Select an image that you think best matches the caption. Your output must follows the following JSON format:
\begin{lstlisting}
```json
{
    "thought": "Your thought here.",
    "choice": "Write a single integer representing the index of the card you selected. Valid choices are {{valid_choices}}.",
}
```
\end{lstlisting}
\end{prompt}

\end{document}